\pdfoutput=1
\documentclass[11pt]{article}

\usepackage[]{naacl2021}

\usepackage{times}
\usepackage{latexsym}

\usepackage{hyperref}
\usepackage{url}
\usepackage{graphicx}
\usepackage{multirow}
\usepackage{tabularx}

\usepackage[T1]{fontenc}
\usepackage[utf8]{inputenc}

\usepackage{microtype}

\usepackage{amsmath,amsfonts,bm}

\def\eqref#1{equation~\ref{#1}}
\def\1{\bm{1}}

\DeclareMathAlphabet{\mathsfit}{\encodingdefault}{\sfdefault}{m}{sl}
\SetMathAlphabet{\mathsfit}{bold}{\encodingdefault}{\sfdefault}{bx}{n}

\title{Iterative Decoding for Compositional Generalization in Transformers}

\author{Luana Ruiz \\
    ESE Department\\
    University of Pennsylvania\\
    \texttt{rubruiz@seas.upenn.edu} \\\And
    Joshua Ainslie \and Santiago Onta{\~n}{\'o}n \\
    Google Research \\
    \texttt{\{jainslie,santiontanon\}@google.com} \\}

\begin{document}
\maketitle
\begin{abstract}
Deep learning models generalize well to in-distribution data but struggle to generalize compositionally, i.e., to combine a set of learned primitives to solve more complex tasks. In sequence-to-sequence (seq2seq) learning, transformers are often unable to predict correct outputs for longer examples than those seen at training. This paper introduces {\em iterative decoding}, an alternative to seq2seq that (i) improves transformer compositional generalization in the PCFG and Cartesian product datasets and (ii) evidences that, in these datasets, seq2seq transformers do not learn iterations that are not unrolled. In iterative decoding, training examples are broken down into a sequence of intermediate steps that the transformer learns iteratively. At inference time, the intermediate outputs are fed back to the transformer as intermediate inputs until an end-of-iteration token is predicted. %We show that transfomers trained via iterative decoding outperform seq2seq on PCFG, and solve Cartesian products between vectors longer than those seen at training---a task at which seq2seq models fail---with 100\% accuracy.
We conclude by illustrating some limitations of iterative decoding in the CFQ dataset.
\end{abstract}

\section{Introduction} 
\label{sec:intro}

Deep learning architectures achieve state-of-the-art (SOTA) results in a wide array of machine learning problems, where their impressive performance is attributed to their ability to generalize \citep{goodfellow2016deep,lecun2015deep}. However, this ability is typically limited to generalization under the statistical learning paradigm, i.e., in-distribution generalization, and does not encompass generalizing compositionally. Compositional generalization is the ability of a model to combine a set of learned primitives to execute more complex tasks. For instance, for a ground robot whose motion planner has learned to execute the instructions ``walk'', ``jump'', and ``jump right'', generalizing compositionally would be to be able to execute the instruction ``walk right'' \citep{lake2018generalization}.

In machine learning, compositional generalization is desirable for two reasons. First, because it is a crucial aspect of intelligence observed in both humans and classical artificial intelligence. In humans, a prevailing example is the way children can solve complicated mathematical expressions after being taught basic arithmetics. Second, because it can increase a model's data efficiency. By endowing models with the ability to extrapolate to unseen examples involving compositions of primitives not seen during training, compositionality acts as a mechanism for data augmentation.

\begin{figure*}[ht]
\begin{center}
\includegraphics[width=12cm]{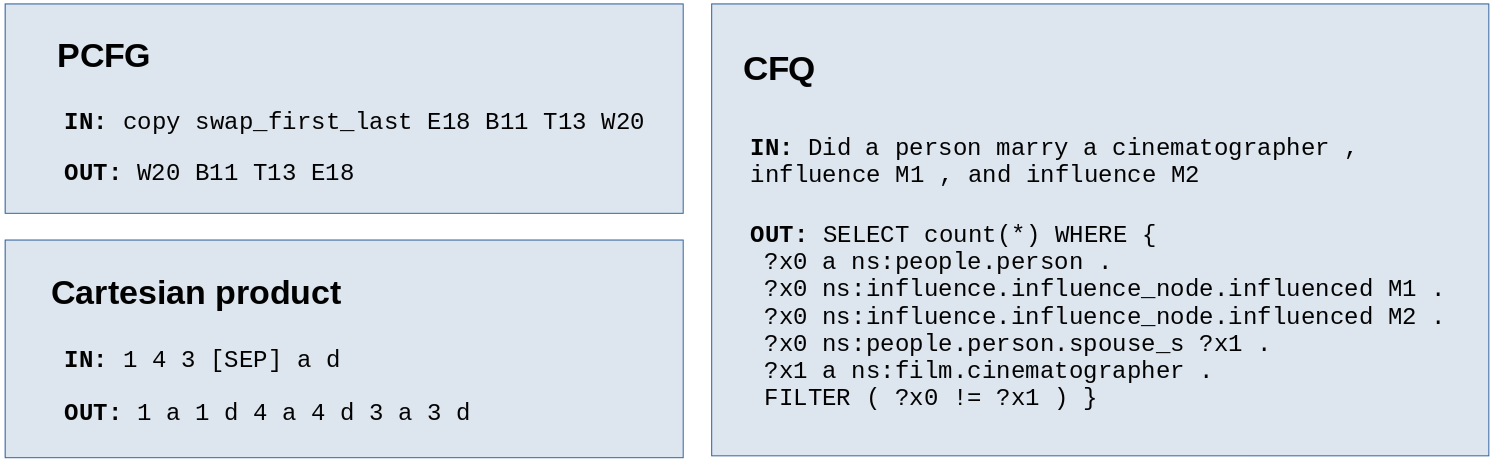}
\end{center}
\caption{Examples of input-output pairs in the PCFG, Cartesian product, and CFQ datasets.}
\label{fig:data_examples}
\end{figure*}

In this paper, our goal is to increase compositional generalization in transformers with particular focus on natural language-like tasks, where compositionality is key. 
Understanding that for models to execute composite tasks they need to be taught \textit{how to compose}, we introduce {\em iterative decoding}, an alternative to sequence-to-sequence (seq2seq) learning that decomposes the process of mapping the inputs to the outputs of each example into a sequence of intermediate steps which the transformer learns to perform iteratively. During training, each input-output pair is converted into a sequence of ``intermediate input-intermediate output'' pairs. During prediction, the predicted intermediate outputs are adapted into the subsequent intermediate inputs, which are fed back to the transformer until an end-of-iteration (EOI) token is produced.

Our main contributions are (i) showing that iterative decoding, especially when combined with architectural modifications such as relative attention \citep{shaw2018self} and copy decoding \citep{gu2016incorporating}, improves compositional generalization in transformers trained on PCFG (top left of Fig. \ref{fig:data_examples}) \citep{hupkes2020compositionality} and Cartesian product (bottom left of Fig. \ref{fig:data_examples}), and (ii) evidencing that, in these datasets, seq2seq transformers cannot learn iterations unless they are unrolled. PCFG is a string editting dataset in which we consider two compositionally hard splits. In Cartesian product, the goal is to generalize to longer input vectors than those on which the model is trained. We also present numerical results on CFQ \citep{keysers2019measuring}, a semantic parsing dataset consisting of natural language questions and SPARQL queries (right of Fig. \ref{fig:data_examples}). The results obtained on CFQ evidence a limitation of iterative decoding, which can be sensitive to the ordering of the intermediate steps. 
%This experiment also evidences another general limitation of transformers, which is that they struggle to learn how to sort.

We believe that iterative decoding can potentially improve transformer performance in other compositionally hard problems in NLP. Nevertheless, the definition of the intermediate steps for these tasks is dataset-specific and is out of the scope of this work. Our goal is not to propose iterative decoding as a ``one-size-fits-all'' solution, but rather to show that, when a heuristic is available (as is the case of the datasets we consider), iterative decoding can improve the ability of transformers to generalize compositionally [cf. Remark 1]. In the case of PCFG, Cartesian and CFQ, the specific construction of the intermediate steps is detailed in the experimental results section (Secs. \ref{sbs:iter_pcfg} through \ref{sbs:iter_cfq}). 

%The rest of this paper is organized as follows. Sec. \ref{sec:background} discusses related work on compositional generalization and transformers. Sec. \ref{sec:iterative} introduces iterative decoding. Numerical results on the PCFG, Cartesian product, and CFQ datasets are presented and discussed in Sec. \ref{sec:results}. Sec. \ref{sec:conclusions} presents future research directions and concluding remarks.

\section{Background}
\label{sec:background}

In this section, we introduce some background and related work on compositional generalization and transformer architectures.

\subsection{Compositional generalization}
\label{sbs:compositionality}

Compositional generalization (or compositionality~\cite{fodor2001language}) refers to the ability of a model/agent that has learned to perform a set of basic operations---\textit{primitives}---to generalize to more complex operations, i.e., operations consisting of \textit{compositions} of the learned primitives \citep{lake2018generalization}. Examples of operations requiring compositionality are shown in Fig. \ref{fig:data_examples} for three datasets. For instance, the top-left corner shows an example from the PCFG~\citep{hupkes2020compositionality} dataset. In some versions of this dataset, the model is trained to solve several atomic string editing operations (such as \texttt{copy} and \texttt{swap\_first\_last}), and how to compose them. The test data contains longer sequences with more operations than seen during training. Hence, the model's performance relies on compositional generalization. This string editing example can be seen as an instance of {\em productivity}, one of the five types of compositional generalization identified by \citet{hupkes2020compositionality} which involves generalizing to longer examples. Another type of compositional generalization is {\em systematicity}, which is the ability to recombine known parts and rules in ways different than those seen during training.

Early works on compositionality have explored the limitations of machine learning models in generalizing compositionally. \citet{livska2018memorize} showed that, while it is theoretically possible for a recurrent neural network (RNN) to generalize in this way, only a small fraction of the models they trained behaved compositionally. \citet{lake2018generalization} proposed SCAN, a dataset consisting of navigation commands to be mapped to action sequences, and observed that while RNNs trained on it generalized well when the differences between the training and test sets were small, they failed when more systematic compositional skills were required. Other datasets created to measure compositionality include ListOps \citep{nangia2018listops}, where latent tree models perform worse than purely sequential RNNs, and PCFG \citep{hupkes2020compositionality} and CFQ \citep{keysers2019measuring}, where neither long-short term memory (LSTM) nor transformer-based architectures perform well.

\begin{figure*}[t]
\begin{center}
\includegraphics[width=13cm]{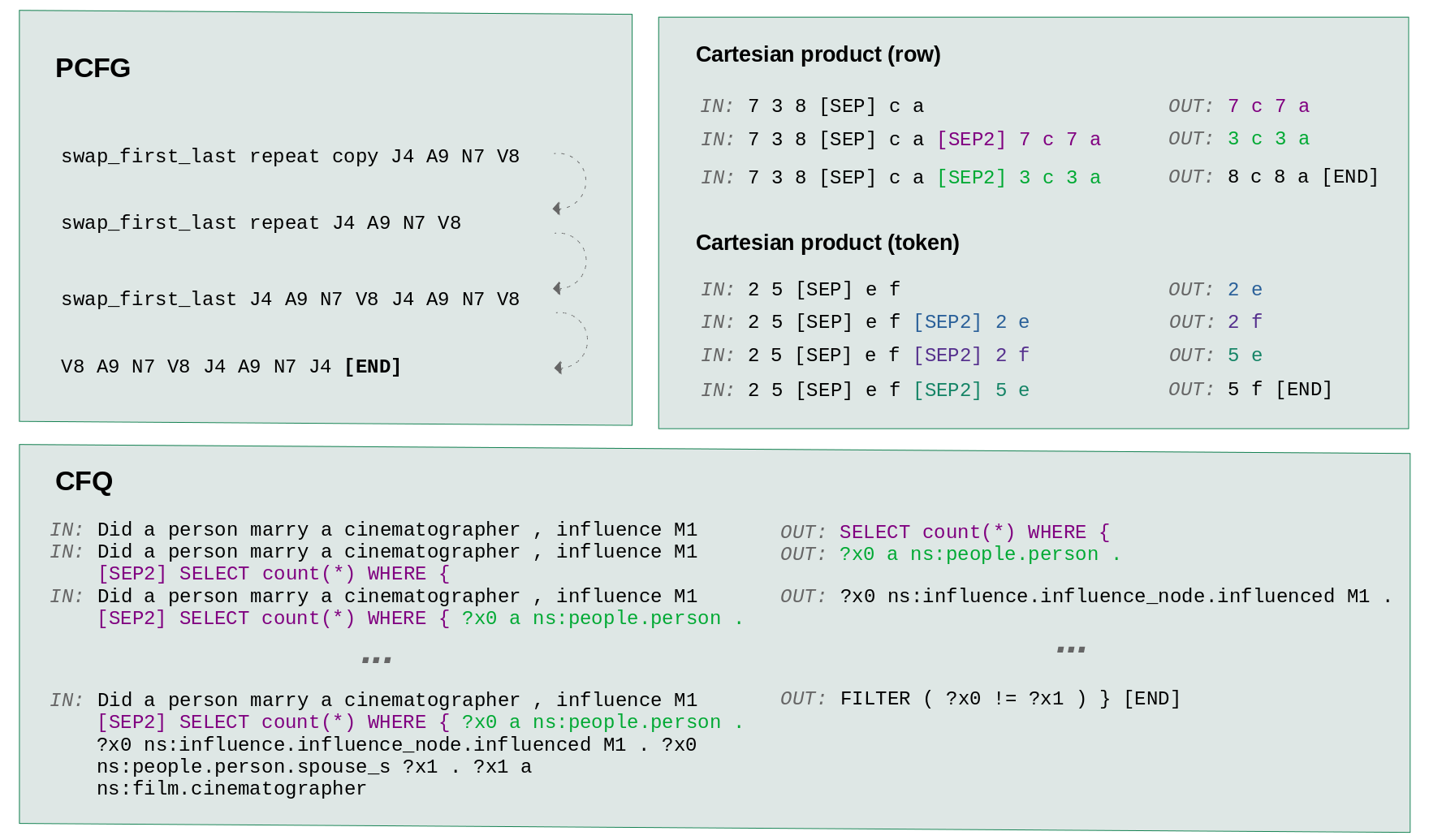}
\end{center}
\caption{Examples of intermediate input-output pairs in the PCFG and Cartesian product datasets.}
\label{fig:it_dec_examples}
\end{figure*}

More recently, a popular research direction is to try to endow these machine learning models with a ``compositional generalization bias''. \citet{kim2021improving} saw benefits in converting CFQ into a classification task and using structural annotations (e.g., entity links) as attention masks in transformers. \citet{ontanon2021making} were able to improve transformer compositional generalization on a variety of compositionally hard datasets by making architectural modifications such as relative attention, copy decoders, and weight sharing. Taking a similar approach, \citet{csordas2021devil} observed performance improvements from relative position encodings and scaled embeddings. Other strategies to improve compositional generalization include increased pretraining \citep{furrer2020compositional}, data augmentation \citep{andreas2019good} and differentiable neurocomputers \citep{graves2016hybrid}.

Closely related to our work, PonderNet \citep{banino2021pondernet} trains a model that iterates internally 
%adaptively decomposes tasks into computational steps 
to achieve a better compromise between training accuracy and generalization. It predicts both an output and a halting probability at each step, operating recurrently. Iterative decoding also operates recurrently, but with two important differences: (i) the intermediate steps are supervised, and (ii) the model is trained to produce a special token to indicate the end of the iteration rather than predicting a halting probability at each step.
%Another difference with our approach is that while PonderNet aims to learn which intermediate steps to perform, in iterative decoding, we provide supervision to these intermediate steps.

%\red{Maybe write a bit about how compositional generalization is a type of out-of-domain generalization? [santi: if you add this, I'd add this at the end of the first paragraph]}

\subsection{Transformer model}
\label{sbs:transformer}

In this paper we focus on transformer models. Despite struggling to generalize compositionally, transformer-based architectures such as BERT \citep{devlin2018bert} and T5 \citep{raffel2019exploring} were popularized by their remarkable performance in machine translation \citep{zhu2020incorporating}, question answering \citep{ainslie2020etc}, summarization \citep{zhang2019hibert} and other natural language processing (NLP) tasks.  

Introduced by \citet{vaswani2017attention}, the basic transformer model is composed of an encoder and a decoder. The encoder is made up of layers consisting of a self-attention sublayer and a feedforward sublayer. The decoder has the same structure, but with an additional attention sublayer to compute the decoder-to-encoder attention. The input to the transformer is a sequence of token embeddings. Since these embeddings do not carry information about the position of each token in the sequence, a position encoding is typically added to the input embeddings. These are then fed to the encoder, which encodes all tokens at once and forwards the result to the decoder. From the encoded input and the decoded output tokens generated so far, the decoder generates the distribution of the next output token, one token at a time. 

We experiment with two extensions of the original transformer architecture: \textbf{relative position encodings} \citep{shaw2018self} and \textbf{copy decoding} \citep{gu2016incorporating}. We chose these techniques because they have been shown to improve compositionality in seq2seq learning \citep{ontanon2021making} and require less parameters than larger transformer models. To each pair of tokens in the input, relative position encodings assign a label that equal to the minimum between their relative distance and a relative attention radius. Relative position encodings are thus position invariant, which means that two tokens that are $k$ positions apart will attend to each other in the same way regardless of their absolute positions in the sequence. This also makes them invariant to length of the sequence which, intuitively, should improve compositional generalization.

Copy decoding involves adding a learnable parameter that allows to switch between the decoder and a copy decoder which produces an independent embedding that can be interpreted as a ``copy'' from the input sequence.
This helps with compositional generalization because many tasks, e.g., PCFG, have a type of input-output symmetry that requires producing parts of the input at the output.
Copy decoding can also be seen as ``pointing'' to tokens in the input sequence. This pointing mechanism was proposed in \citep{gulcehre2016pointing}, and \citet{oriol2015pointer} showed that it  allows models to generalize beyond the lengths they are trained on. 
%can directly copy tokens from the input sequence.

%\subsection{Datasets}
%\label{sbs:datasets}

%We consider three datasets---PCFG, Cartesian product, and CFQ---which are discussed in detail in the following sections.

%\subsubsection{PCFG}

%\subsubsection{Cartesian product} \label{sbs:Cartesian}

%\subsubsection{CFQ}

\section{Iterative Decoding}
\label{sec:iterative}

To improve compositional generalization in transformers, we introduce iterative decoding. As illustrated on the right hand side of Fig. \ref{fig:seq2seq_vs_it_dec}, iterative decoding consists of predicting a series of intermediate outputs $y_1, y_2, y_3, \ldots$ from an input $x=x_0$, and then adapting these outputs into intermediate inputs $x_i=y_i$, $i>0$, that are fed back to the model until the final output $y_N=y$ is predicted.
This can be visualized by considering the PCFG example $x=$ ``\texttt{swap\_first\_last repeat copy J4 A9 N7 V8}'' on the left hand side of Fig. \ref{fig:it_dec_examples}. A seq2seq transformer trained on the PCFG dataset is expected to output $y=$ ``\texttt{V8 A9 N7 V8 J4 A9 N7 J4 [END]}'' in one forward pass (i.e, to go from top to bottom in the Fig.). However, in iterative decoding, the transformer's output to the input $x_0=x$ would be $y_1=$ ``\texttt{swap\_first\_last repeat J4 A9 N7 V8}'', which is the first intermediate output of iterative decoding (the second string from the top), and corresponds to just executing one of the operations in the input, \texttt{copy}. Setting $x_1=y_1$ and feeding this instruction back to the transformer, we obtain $y_2$= ``\texttt{swap\_first\_last J4 A9 N7 V8 J4 A9 N7 V8}'' (the third string from the top). The intermediate output $y_2$ then becomes the intermediate input $x_2$, which the transformer processes to produce the final output $y_3=$ ``\texttt{V8 A9 N7 V8 J4 A9 N7 J4 [END]}''.

The main motivation for iterative decoding comes from the very idea of compositional generalization: by decomposing complex instructions into intermediate steps, iterative decoding essentially teaches models how to compose. Another motivation, related to the first, is that iterative decoding mimics how humans are taught how to perform many compositional tasks. For example, when teaching how to solve the arithmetic expression $2\times(1+1)$, first we demonstrate how to solve the inner sum, then how to eliminate the parentheses, and finally how to compute the product. A third motivation for iterative decoding is that learning step by step can potentially prevent the model from learning shortcuts, i.e., from overfitting to specific examples instead of learning compositional symmetries applicable to examples requiring the same type of compositional generalization to be solved.
%one of the biggest obstacles to compositional generalization in seq2seq, because it reduces the difficulty of the tasks that the transformer needs to learn to execute in a single forward pass.
%, which is one of the biggest obstacles to compositional generalization in seq2seq.

\subsection{Implementation}

To implement iterative decoding, we modify both how models are trained and how they predict. 

\begin{figure}[t]
%\begin{center}
\includegraphics[width=7.5cm]{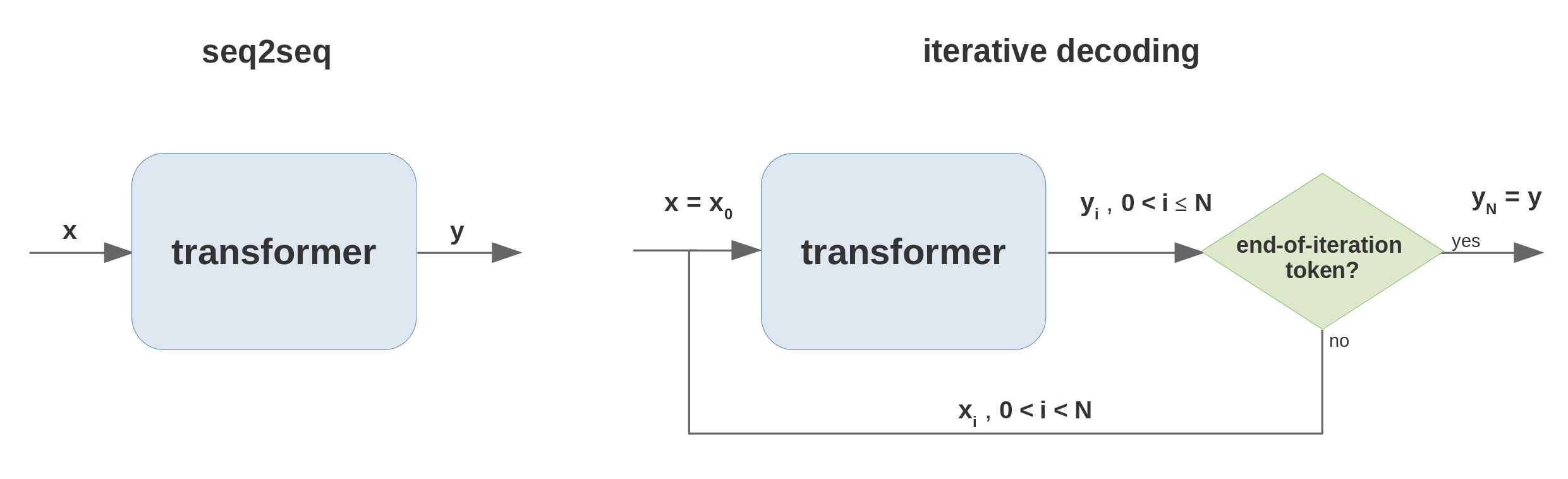}
%\end{center}
\caption{Prediction routines for the seq2seq transformer (left) and iterative decoding transformer (right).}
\label{fig:seq2seq_vs_it_dec}
\end{figure}

\begin{table*}[t]
\caption{Sentence accuracy on the training and test sets for the random, productivity and systematicity splits of the PCFG dataset, for seq2seq and iterative decoding.}
\label{pcfg-table}
{\small
\begin{center}
\begin{tabular}{lcccccc}
\hline
\multirow{2}{*}{\textbf{Model}} &\multicolumn{2}{c}{\bf Random} &\multicolumn{2}{c}{\bf Productivity} &\multicolumn{2}{c}{\bf Systematicity} \\
 &train &test &train &test &train &test \\
\hline
Seq2seq             &87.2\% &85.9\% &90.6\% &34.2\% &89.5\% &64.8\% \\
+ Relative attention            &98.1\% &97.4\% &98.8\% &65.1\% &98.2\% &85.7\%\\
+ Copy decoder      &97.7\% &97.0\% &98.3\% &63.9\% &98.2\% &85.1\%\\
\hline
Iterative decoding            &96.5\% &93.2\% &96.6\% &45.7\% &94.6\% &82.9\% \\
+ Relative attention             &99.8\% &99.2\% &100\% &91.9\% &99.7\% &97.0\%\\
+ Copy decoder      &99.7\% &\textbf{99.4\%} &100\% &\textbf{93.3}\% &99.8\% &\textbf{97.8}\%\\ \hline
\end{tabular}
\end{center}
}
\end{table*}

\noindent \textbf{Training.} Instead of being trained on the original inputs and outputs, iterative decoding transformers are trained on the ``intermediate input-intermediate output'' pairs $(x_{i-1},y_i)$ for $1 \leq i \leq N$. These inputs and outputs are pre-generated, and their form is specific to each task (see Sec. \ref{sec:results} for examples for PCFG, Cartesian product, and CFQ). Naturally, the addition of the ``intermediate input-intermediate output'' pairs to the training data increases the number of training samples, which leads to an increase of the computational cost per epoch proportional to the average number of intermediate steps. In our experiments, we avoid this by training the iterative decoding and the seq2seq transformer not for the same number of epochs, but for the same number of steps.
Another difference with seq2seq training is that in iterative decoding an EOI token has to be added to the training outputs so that the model learns when to stop. Intuitively, the intermediate outputs correspond to providing supervision on the steps that the transformer should execute to solve the task. This is analogous to how, rather than learning to solve arithmetic expressions by looking at input-output mappings, humans are taught to solve the intermediate steps involved in their computation. Further note that there is no recurrence involved in the training of the iterative decoding transformer. An intermediate output produced by the model during training does not need to be fed back to the model as an intermediate input, because the subsequent intermediate input is already an input sample of the training set. Moreover, ``intermediate input-intermediate output pairs'' corresponding to the same example do not need to be presented to the transformer in any particular order, and can be shuffled at random in the training set.

\noindent \textbf{Prediction.} Depending on the number of intermediate steps necessary to iteratively decode an example, the prediction requires multiple forward passes of the transformer. Hence, it is implemented as a while loop where the stopping condition is finding the EOI token. This is illustrated on Fig.~\ref{fig:seq2seq_vs_it_dec}, which compares seq2seq (left) with iterative decoding predictions (right). After each intermediate prediction, a data processing step may be needed to adapt the intermediate outputs into the following intermediate inputs in some datasets. As we could see from the example above, this is not necessary in PCFG, because it has a built-in recursive structure. But it is necessary in Cartesian product and CFQ as we detail in Sec. \ref{sec:results}. While ideally this processing step should be learned, in this paper we provide it manually to develop a preliminary understanding of the advantages of iterative decoding.

\noindent \textbf{Remark 1 (Construction of intermediate steps.)} The construction of the intermediate steps is dataset-specific. In PCFG, Cartesian and CFQ, this construction is detailed in Secs. \ref{sbs:iter_pcfg} through \ref{sbs:iter_cfq}. While in principle the intermediate steps do not have to fit any specific requirements, their definition will typically rely on a heuristic method to decompose the input into ``intermediate input-intermediate output'' pairs, which push the model to learn the basic primitives operations in a dataset, which can then be composed to perform more complex operations. In some datasets, e.g., PCFG, this is straightforward. In others, e.g., Cartesian, there are multiple admissible options to decompose the input into intermediate steps. In more complex problems such as CFQ and other semantic parsing tasks, this decomposition is less obvious and requires some engineering. The definition of heuristics for different datasets is out of the scope of this paper as our goal is not to propose iterative decoding as a ``one-size-fits-all'' solution, but rather to show that, when a heuristic is available, iterative decoding can improve the ability of transformers to generalize in a compositional way.

\section{Results and Discussion}
\label{sec:results}

In this section we describe the iterative decoding schemes for PCFG, Cartesian product and CFQ, and present and discuss numerical results obtained for seq2seq and iterative decoding transformers on these datasets. All transformers have $\ell = 6$ encoder/decoder layers, embedding dimension $d = 64$, feedforward dimension $f = 256$ and $h = 4$ attention heads. For each dataset, the seq2seq and the iterative decoding transformer are trained for the same number of training steps and each experiment is repeated $3$ times. Implementation details can be found in Appendix \ref{app:implementation}.

%\red{Results go here. Include examples of mistakes in the seq2seq and iterative decoding scenarios.}

\subsection{PCFG}
\label{sbs:iter_pcfg}

PCFG is an artificial translation dataset proposed by \citet{hupkes2020compositionality} and generated by a probabilistic context free grammar. A more detailed description can be found in Appendix \ref{app:addtl_datasets}, and an example of input-output pair is shown on the top left corner of Fig. \ref{fig:data_examples}.
There are six training-test splits of PCFG. The first is a random split which we use as a baseline. The other five are compositionally hard splits used to measure the five different types compositional generalization. We focus on two of them: productivity and systematicity. In the productivity split, the training samples have up to 8 string operations, while the test samples have 9 or more. In the systematicity split, the operations in the test set they are combined in different ways than in the training set. 
%Fig. \red{X} shows examples of training and test samples in each split.

We apply iterative decoding to PCFG by breaking down each example into a number of intermediate steps equal to the number of string editing operations present in the original input. Each intermediate step solves the rightmost instruction in the current intermediate input. Up until the last step, all of the intermediate outputs are themselves string editing instructions. Hence, the intermediate outputs do not need to be adapted and serve as the intermediate inputs to the next step. Hence, the only additional processing of the dataset needed for iterative decoding is the addition of the EOI token \texttt{[END]} at the end of the final output. An iterative decoding PCFG example with three intermediate steps is shown on the left hand side of Fig. \ref{fig:it_dec_examples}.

To compare iterative decoding with seq2seq, we start with a basic transformer model with absolute position encodings as in the original architecture by \citet{vaswani2017attention}. This is so we can observe the advantages of iterative decoding in the absence of other compositional generalization biases. The results are reported in Table \ref{pcfg-table}, where we see that in the random split of the data, the test accuracy of the seq2seq model is close to its training accuracy. This indicates that the model generalizes well to in-distribution samples. In contrast, in the productivity and systematicity splits there is a dramatic drop in test accuracy. Hence, the basic seq2seq transformer struggles to generalize compositionally. In the fourth row of Table \ref{pcfg-table}, we see that for the iterative decoding transformer the gap between training and test accuracy is much smaller. This indicates that iterative decoding increases the ability of transformers to generalize compositionally.

Although iterative decoding helps with compositionality, the iterative decoding test accuracy is still low compared to their training accuracy. This implies that composing individual operations into complex instructions is only one facet of compositionality, which makes sense as decomposing complex instructions into individual operations only helps if the model can execute each operation correctly (see Appendix \ref{app:addtl_results} for more details). Therefore, we repeat our experiments with transformers including modifications shown to increase compositional generalization in seq2seq learning \citep{ontanon2021making}: \textit{relative attention} \citep{shaw2018self} and \textit{copy decoding} \citep{gu2016incorporating}.

The results for seq2seq and iterative decoding with relative attention are shown in the second and fifth rows of Table \ref{pcfg-table}. The relative attention radius is $r=8$. As expected, relative attention helps both models with compositionality, particularly in the productivity split. Moreover, iterative decoding achieves a much better test accuracy, of over 90\%, on both the productivity and systematicity splits. Results for transformers with relative attention and copy decoders are shown in the third and sixth rows of Table \ref{pcfg-table}. Additionally, examples of errors made by both transformers can be found in Appendix \ref{app:error_examples}. While adding a copy decoder does not improve compositional generalization in the seq2seq transformer, it helps in iterative decoding, nearly closing the gap between training and test accuracy on the systematicity split, and leading to a ~2\% increase in test accuracy on the productivity split---which can be attributed to the ability to copy the longer strings in this split.

\begin{table*}[tb]
\caption{Sentence accuracy achieved by the seq2seq and iterative decoding transformers with relative attention ($r=8$) on the training set and on multiple test sets of the Cartesian product dataset.}
{\small
\label{cart-table}
\begin{center}
\begin{tabular}{lccccc}
\hline
\multicolumn{2}{c}{} &\multicolumn{4}{c}{\bf Iterative decoding} \\
\multirow{1}{*}{\textbf{Split}} &\multirow{1}{*}{\textbf{Seq2seq}} &\multicolumn{2}{c}{short inputs} &\multicolumn{2}{c}{long inputs}\\
\multicolumn{2}{c}{} &row &token &row &token \\
\hline
Train (up to 5 numbers/letters)       &100\%      &100\% &100\% &100\% &100\%\\
Test  (up to 5 numbers/letters)       &97.8\%     &100\% &100\% &100\% &100\%\\
\hline
Test  (6 numbers, 5 letters)    &14.3\%     &89.2\% &\textbf{100\%} &\textbf{100\%} &\textbf{100\%}\\
Test  (5 numbers, 6 letters)    &12.2\%     &0\% &99.5\% &0\% &\textbf{100\%}\\
\hline
Test (6 numbers/letters)        &1.1\%      &0\% &98.7\% &0\% &\textbf{100\%}\\ \hline
\end{tabular}
\end{center}
}
\end{table*}

\subsection{Cartesian product}
\label{sbs:iter_cart}

In the Cartesian product dataset  \citep{ontanon2021making}, the inputs are two vectors and the outputs are their Cartesian product. A more detailed description of this dataset can be found in Appendix \ref{app:addtl_datasets} and an example of input-output pair is shown on the bottom left corner of Fig. \ref{fig:data_examples}.
We consider four train-test splits. In the first split, both the training and test set consists of samples with up to five numbers and letters drawn i.i.d and split at random. This is the ``easy split''. In the other splits, the training set is the same as in the first split, but the test set consists of examples with either more number, more letters, or both. These are ``hard splits'', which we use to measure productivity. 

To iteratively decode a Cartesian product, we first need to define what are going to be the iterative decoding intermediate steps. 
%There are many possibilities; we could, for instance, define an intermediate step as predicting one output token at a time, or a sequence of tokens with fixed length. 
We consider two options as illustrated on the right corner of Fig. \ref{fig:it_dec_examples}. The first is decoding one \textit{row} at a time., which entails decoding the Cartesian product between one element from the first vector and all elements of the second vector at each intermediate step. The second is decoding one \textit{token pair} at a time, which entails decoding only the product between one element of the first vector and one element of the second vector at each intermediate step. If the lengths of the input vectors are $\ell_1$ and $\ell_2$ respectively, decoding row by row requires $\ell_1$ and token by token $\ell_1 \times \ell_2$ intermediate steps. 

When decoding rows, the intermediate output at a given step is the current row. When decoding token pairs, the intermediate outputs are the current token pairs. Since these intermediate outputs do not carry any information about the next row or pair of tokens, they cannot be used as intermediate inputs. To construct intermediate inputs, we concatenate a copy of the original input---the two vectors separated by the \texttt{[SEP]} token---with a second separation token \texttt{[SEP2]} followed by either (i) the last intermediate output, or (ii) all the intermediate outputs so far. Scenario (i), which is illustrated on the right hand side of Fig. \ref{fig:it_dec_examples}, yields \textit{short intermediate inputs} where the last intermediate output acts as a ``pointer'' to where the decoding process stopped in the previous step. Scenario (ii) produces \textit{long intermediate inputs}.
While in both scenarios the intermediate outputs need to be concatenated to produce the final prediction, their prediction routines are different because scenario (i) only needs to append the current intermediate output to the input vectors to produce the next intermediate output, but scenario (ii) needs to append the current intermediate output to the last intermediate input to produce the next intermediate input.
%Note that there are many other possibilities to construct intermediate inputs (e.g., only including the last two, three, four, etc. intermediate outputs). There are also different options for the intermediate outputs. Instead of being only the next row or pair of tokens, for instance, they could be all the rows or all the token pairs so far.

To analyze the compositional generalization of seq2seq and iterative decoding transformers on the Cartesian product dataset, we consider the following experimental setup. Both transformers are trained on samples with up to five numbers and letters. Then, they are tested on the four different test sets described above: up to five numbers and letters; six numbers and five letters; five numbers and six letters; and six numbers and letters. The second, third and fourth test sets can be seen productivity tests. 
Additionally, we only report results for transformers with relative attention (relative radius $r=8$) as they were the best performing architecture in our experiments.

The average training and test accuracies achieved by the seq2seq model, as well as by the iterative decoding model in the short/long intermediate input and row/token scenarios, are reported in Table~\ref{cart-table}. 
The seq2seq model (first column) achieves 100\% accuracy on the training set and close to that on the ``easy'' test set with up to five numbers and letters. However, it pretty much fails in all of the ``hard'' test sets, implying that it cannot generalize compositionally when even one extra token is added to the input. 
The models that decode one row at a time (second and fourth columns) do slightly better as they achieve accuracy closer to the training accuracy in the test set with six numbers and five letters. However, they still fail at the test sets with six letters, which means that the iterative decoding transformer trained to decode one row at a time only generalizes well to calculating Cartesian products with a larger number of numbers or, equivalently, of rows. 

In contrast, the iterative decoding models that decode one pair of tokens at a time (third and fifth columns) achieve close to 100\% accuracy in all compositionally hard splits.
This means that the iterative decoding transformer can only generalize to longer iterations when these iterations are the ones that were unrolled during training via iterative decoding. If we take a transformer that has been trained to decode rows, and add one more letter to the input---resulting in one more pair of tokens in each row---it will not be able to predict the extra pair because it has learned how to unroll more rows through iterative decoding but not more token pairs within a row. We thus conclude that in the Cartesian product dataset transformers struggle to learn iteration by themselves, i.e., without the help of iterative decoding. This is an important result because, despite being universal function approximators in theory~\citep{yun2019transformers}, it sheds light onto what transformers can actually learn in practice. Finally, the difference between long and short inputs is not substantial, but longer inputs seem to be better, probably because they provide the transformer with more memory (i.e., more information about the previous intermediate steps).

\subsection{CFQ}
\label{sbs:iter_cfq}

Introduced by \citet{keysers2019measuring}, the CFQ dataset consists of natural language questions and their corresponding SPARQL queries against the Freebase knowledge base. \citet{keysers2019measuring} introduces a number of compositionally hard splits of the CFQ dataset. In this paper, we focus on the MCD1 split. Additional details are described in Appendix \ref{app:addtl_datasets} and an example of question and query are shown on the right hand side of Fig. \ref{fig:data_examples}.

Both PCFG and Cartesian product were well adapted for iterative decoding because, in PCFG, the recursive structure of the inputs makes it easy to define the intermediate steps, and in Cartesian product we have the flexibility to choose their granularity. On the CFQ dataset, defining the intermediate steps is less obvious. The natural choice is to define each intermediate output as a clause of the query as illustrated in Fig. \ref{fig:it_dec_examples}. However, unlike in PCFG---where the order of the intermediate steps was defined by the recursion---and in Cartesian product---where the order of the tokens in the input determines the order of the intermediate steps--- this ordering of the intermediate steps is not very ``natural'' because it is alphabetic. Hence, on CFQ transformers also have to learn how to sort. 

To make learning this ordering easier for the transformer, we define long intermediate inputs for the intermediate steps. These intermediate inputs are constructed by concatenating the question with all of the previous intermediate outputs so far. As such, on CFQ the iterative decoding prediction routine is the same as for Cartesian product with long inputs: we append the current intermediate output to the previous intermediate input to obtain the next intermediate input, and, once the EOI token is predicted, concatenate all of the intermediate outputs to obtain the query prediction.

\begin{table}[tb]
{\small
\caption{Sentence accuracy achieved by the seq2seq and iterative decoding transformers with relative attention ($r=8$) on the training and test sets of the MCD1 split of the CFQ dataset.}
\label{cfq-table}
\begin{center}
\begin{tabular}{lcc}
\hline
\textbf{Split} &\textbf{Seq2seq} &{\bf It. decoding}\\
\hline
Train     &99.8\%      &99.7\% \\
Test     &37.1\%      &32.5\% \\ \hline
\end{tabular}
\end{center}
}
\end{table}

The average training and test accuracies are shown in Table \ref{cfq-table}, where we only report results for  the best performing model in our experiments---relative attention with relative radius $r=8$. While the accuracies achieved by our models are lower than the SOTA results reported in \citep{furrer2020compositional} with pretraining, note that our goal is not to achieve the best possible performance for CFQ, but rather to compare transformers with the same architecture trained via seq2seq and iterative decoding. Both the seq2seq and the iterative decoding transformer exhibit low compositionality, however, iterative decoding performs worse than seq2seq. This reinforces the limitations of iterative decoding that we observed in Cartesian product, namely, that iterative decoding performance is largely dependent on how we define the intermediate steps.

In CFQ specifically, we also hypothesize that the worse performance of iterative decoding is tied to the alphabetical ordering of the clauses, as it does not follow naturally from the grammatical structure of the input. Even though sorting these clauses is something that both the seq2seq and the iterative decoding transformer have to learn how to do, in iterative decoding the transformer has to sort at all intermediate steps, so there are more opportunities to make mistakes. In other words, the error probability is larger in iterative decoding, because it compounds with each intermediate step. 

\section{Conclusions}
\label{sec:conclusions}

This paper introduces iterative decoding as an alternative to seq2seq learning. Through numerical experiments on PCFG and Cartesian product, we demonstrate that, in general, seq2seq transformers do not learn iterations that are not unrolled. By unrolling them, iterative decoding improves transformer compositional generalization. However, iterative decoding has a limitation, which is that it depends on how the intermediate steps are defined. We hypothesize that their ordering is the reason why the seq2seq transformer outperforms iterative decoding on CFQ, as unnatural orderings require transformers to learn how to sort and iterative decoding may increase the overall sorting error probability.
In our future work, we aim to apply iterative decoding strategies to more datasets and understand whether the number of iterative steps can be traded for transformer depth. We will further use iterative decoding to investigate the aspects of compositional generalization that transformers can and cannot learn. A next step is understanding the effect of the order of the intermediate steps and what that says about transformers' ability to sort.
%\subsubsection*{Acknowledgments}

%\vfill

%\pagebreak

\bibliography{iclr2022_conference.bib}
\bibliographystyle{acl_natbib}

\appendix

\begin{figure*}[t]
\begin{center}
\includegraphics[width=6cm]{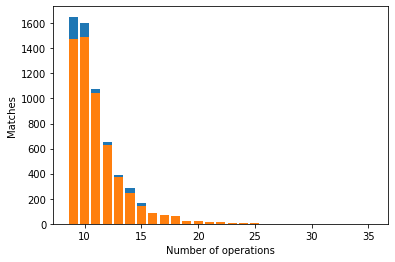}
\hspace{1cm}
\includegraphics[width=6cm]{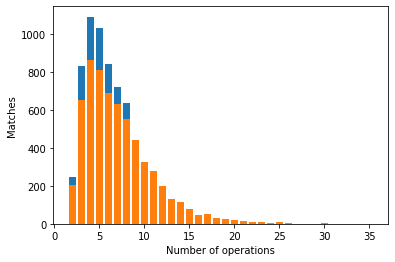}
\end{center}
\caption{Number of correct predictions versus number of string operations in the input for seq2seq (orange) and iterative decoding (blue), on the productivity (left) and systematicity (right) splits of PCFG.}
\label{fig:matches_per_op}
\end{figure*}

\begin{table*}[t]
{\small
\caption{Vocabulary size, training and test samples, and number of training steps for all seq2seq and iterative decoding datasets.}
\label{dataset_stats}
\begin{center}
\begin{tabular}{cccc|ccc|c}
\hline
 &\multicolumn{3}{c|}{\bf Seq2seq} &\multicolumn{3}{c|}{\bf Iterative decoding} &  \\
 &vocabulary &train &test &vocabulary &train &test &steps \\
\hline
PCFG-i.i.d. &534 &82662 &9721 &535 &426558 &9721 &33325 \\
PCFG-prod.  &534 &81010 &11333 &535 &346222 &11333 &27049 \\
PCFG-syst.  &534 &82168 &10175 &535 &403808 &10175 &31458 \\
Cartesian-row  &26 &200000 &1024 &28 &600036 &1024 &4688 \\
Cartesian-token &26 &200000 &1024 &28 &1801869 &1024 &14077 \\
CFQ &181 &95743 &11968 &186 &682470 &11968 &53318 \\ \hline
\end{tabular}
\end{center}
}
\end{table*}

\section{Dataset Details} \label{app:addtl_datasets}

\subsection{PCFG}

 In all splits of the PCFG dataset, the input data consists of string editing instructions with four types of tokens: unary operation tokens (e.g., \texttt{reverse}), binary operation tokens (e.g., \texttt{append}), a string separation token ``\texttt{,}'' (to separate arguments of binary operations), and string elements (e.g., \texttt{B10}, \texttt{D2}). The output data consists of the strings resulting from the application of the operations; see the top left corner of Fig. \ref{fig:data_examples} for an example.
 
 The main challenge of the PCFG dataset is that it requires learning ten string editing operations, some of which are very similar. The unary operation \texttt{echo}, for instance, only differs from \texttt{copy} by repeating the last element of the string. While transformers generally achieve good performance in the random split of the PCFG dataset, the productivity and systematicity splits are harder because transformers tend to learn mappings which do not exploit compositional symmetries.
In particular, a key difficulty of the productivity split is that the model needs to learn to do "recursion" and apply an arbitrary number of operations when input examples grow in length. In some cases, the strings to modify can also be very long, which places an additional capacity burden on transformers by requiring them to learn how to copy strings.

\subsection{Cartesian Product}

 In the Cartesian product dataset, the input consists of two vectors. The first is a vector of numbers. The second is a vector of letters, separated from the numbers by the special token \texttt{[SEP]}. Both numbers and letters are picked at random, without repetition, from the decimal digits and the first ten letters of the alphabet respectively. The output is then the Cartesian product between the first and second input vectors. An example of input-output pair is shown on the bottom left corner of Fig. \ref{fig:data_examples}.
 
 The productivity splits of the Cartesian product dataset are remarkably hard; even transformers with some compositional generalization ability in other mathematical datasets have been seen to fail \citep{ontanon2021making}. This is due to the fact that, in order to solve Cartesian products, models need to learn to execute two nested loops. Moreover, the output is quadratic on the size of the inputs. For models that have to learn to predict an end-of-sequence token, extrapolating to longer sequences than those seen during training has been shown to be difficult \citep{newman2020eos}. 

\begin{figure*}[t]
\begin{center}
\includegraphics[width=6cm]{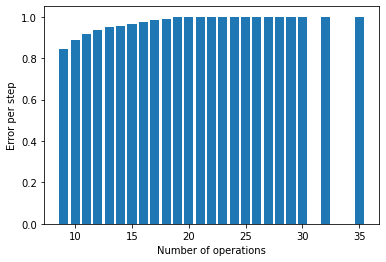}
\hspace{1cm}
\includegraphics[width=6cm]{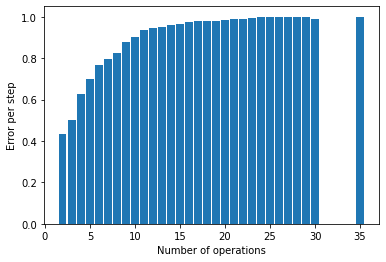}
\end{center}
\caption{Error per intermediate step versus number of string operations in the input for the iterative decoding transformer on the productivity (left) and systematicity (right) splits of PCFG.}
\label{fig:error_per_op}
\end{figure*}

\begin{table*}[t]
{\small
\caption{Productivity error examples for seq2seq and iterative decoding.}
\label{prod-table}
\begin{center}
\begin{tabularx}{\textwidth}{lX}
\multicolumn{2}{c}{\bf Seq2seq} \\
\hline
Input       &  \texttt{shift repeat prepend append Z6 A8 C12 U1 T5 , repeat repeat prepend N8 K15 , S18 B4 , repeat reverse shift echo I2 V2 F5} \\
True output &  \texttt{F5 F5 V2 I2 F5 F5 V2 Z6 A8 C12 U1 T5 S18 B4 N8 K15 S18 B4 N8 K15 S18 B4 N8 K15 S18 B4 N8 K15 I2 F5 F5 V2 I2 F5 F5 V2 Z6 A8 C12 U1 T5 S18 B4 N8 K15 S18 B4 N8 K15 S18 B4 N8 K15 S18 B4 N8 K15 I2}
  \\ 
Prediction  & \texttt{F5 F5 V2 I2 F5 F5 V2 I2 F5 F5 V2 Z6 A8 C12 U1 T5 S18 B4 N8 K15 S18 B4 I2 F5 V2 I2 F5 F5 V2 S18 B4 N8 K15 S18 B4 I2 F5 V2 I2 F5 F5 V2 Z6 A8 C12 U1 T5 S18 B4 N8 K15 S18 B4} \\ \hline
& \\
\multicolumn{2}{c}{\bf Iterative decoding} \\ 
\hline
Input       &  \texttt{remove\_first repeat repeat swap\_first\_last swap\_first\_last R9 Q20 N10 , shift repeat echo repeat V17 V14 E4 A7} \\
True output & \texttt{V14 E4 A7 V17 V14 E4 A7 A7 V17 V14 E4 A7 V17 V14 E4 A7 A7 V17 END} \\ 
Prediction  & \texttt{V14 E4 A7 V17 V14 E4 A7 A7 V17 V14 E4 A7 V17 V14 E4 E4 A7 V17 END} \\ \hline
\end{tabularx}
\end{center}
}
\end{table*}

\subsection{CFQ}

Introduced by \citet{keysers2019measuring}, the CFQ dataset consists of natural language questions and their corresponding SPARQL queries against the Freebase knowledge base. Hence, it can be used to perform semantic parsing by taking the questions as the inputs and the queries as the outputs.
One of the difficulties of CFQ is that some of its examples require solving Cartesian products. As such, in CFQ transformers may face similar challenges to the ones faced in Cartesian product. Another difficulty relates to the ordering of the clauses in the SPARQL query, which are ordered alphabetically by convention. Not only is this ordering different than the one implied by the order of the tokens in the question, it also requires transformers to learn how to sort.

\section{Implementation Details} \label{app:implementation}

Across all experiments, the transformer parameters were the same as in the original implementation in \citep{vaswani2017attention}, including the learning rate schedule. All experiments were run on machines with a single CPU and four Tesla V100 GPUs with batch size $64$ per device.

For each dataset, and for both the seq2seq and iterative decoding splits, the vocabulary size, the size of the training and test sets, and the total number of training steps is shown in Table \ref{dataset_stats}. The iterative decoding vocabularies are larger due to the addition of special start, end and separation tokens. The number of training samples is larger for the iterative decoding splits because they include all intermediate steps. To make for a fair comparison, the number of training steps is the same for iterative decoding and seq2seq.

\begin{table*}[t]
{\small
\caption{Systematicity error examples for seq2seq and iterative decoding.}
\label{syst-table}
\begin{center}
\begin{tabularx}{\textwidth}{lX}
\multicolumn{2}{c}{\bf Seq2seq} \\
\hline
Input       &  \texttt{swap\_first\_last remove\_first F10 E6 T18 , echo append reverse J18 H10 K12 X11 , swap\_first\_last repeat remove\_second copy U4 E15 I2 , X11 C6 W3} \\
True output &  \texttt{U4 K12 H10 J18 I2 E15 I2 U4 E15 U4 X11}
  \\ 
Prediction  & \texttt{U4 K12 H10 J18 I2 E15 I2 U4 E15 U4 U4 E15 X11} \\ \hline
& \\
\multicolumn{2}{c}{\bf Iterative decoding} \\ 
\hline
Input       &  \texttt{repeat remove\_second repeat prepend echo reverse echo prepend Q7 C15 I14 H13 , P9 O5 A12 K19 , remove\_second copy copy G4 W3 U10 S4 , swap\_first\_last echo repeat shift swap\_first\_last I7 S5 Z16 K13 Q9 , copy T16 X18 E15} \\
True output & \texttt{G4 W3 U10 S4 H13 H13 I14 C15 Q7 K19 A12 O5 P9 P9 G4 W3 U10 S4 H13 H13 I14 C15 Q7 K19 A12 O5 P9 P9 G4 W3 U10 S4 H13 H13 I14 C15 Q7 K19 A12 O5 P9 P9 G4 W3 U10 S4 H13 H13 I14 C15 Q7 K19 A12 O5 P9 P9 END} \\ 
Prediction  & \texttt{G4 W3 U10 S4 H13 H13 I14 C15 Q7 K19 A12 O5 P9 P9 G4 W3 U10 S4 H13 H13 I14 C15 Q7 K19 A12 O5 P9 P9 G4 W3 U10 S4 H13 H13 I14 C15 Q7 K19 A12 O5 P9 P9 G4 W3 U10 S4 H13 H13 I14 C15 Q7 K19 A12 O5 P9 K19 END} \\ \hline
\end{tabularx}
\end{center}
}
\end{table*}

\section{Additional Iterative Decoding Results for PCFG} \label{app:addtl_results}

To assess the advantages of iterative decoding  under no other sources of compositional generalization, we consider the base transformer (i.e., without relative attention and copy decoder) and analyze its performance on PCFG per number of string editing operations. Namely, in Fig. \ref{fig:matches_per_op} we plot the number of correct predictions achieved by seq2seq (orange) and iterative decoding (blue) on the productivity and systematicity splits of PCFG. We observe that, in the productivity split, the performance improvement comes mostly from samples with a small number of string editing instructions. Consistent with Table \ref{pcfg-table}, without any other form of compositional generalization bias iterative decoding is more helpful with systematicity. 

We draw a similar conclusion from Fig. \ref{fig:error_per_op}, which plots the error per intermediate step ($(\mbox{test error})^{1/N}$, where $N$ is the number of operations) versus the number of operations in the input for both splits. The error per intermediate step can be seen as the probability of making a mistake at any given intermediate step. On the left, this error approaches one for a smaller number of operations than on the right, indicating that errors compound faster in the productivity split. Interestingly, this Fig. also corroborates our claim from Sec. \ref{sbs:iter_pcfg} that decomposing complex instructions into individual operations only helps if the model can execute each operation correctly. In other words, composing individual operations into complex instructions is only one facet of compositionality, but one with which iterative decoding helps.

\section{Error Examples for PCFG} \label{app:error_examples}

Tables \ref{prod-table} and \ref{syst-table} show examples of wrong predictions made by the seq2seq and iterative decoding transformers with relative attention and copy decoding.

\end{document}